\pdfoutput=1

\documentclass[11pt]{article}

\usepackage[final]{acl}

\usepackage{times}
\usepackage{latexsym}
\usepackage[T1]{fontenc}

\usepackage[utf8]{inputenc}

\usepackage{microtype}
\usepackage{inconsolata}
\usepackage{threeparttable} 

\usepackage{graphicx}

\usepackage[draft]{todonotes}

\title{Spontaneous Speech Variables for Evaluating LLMs' Cognitive Plausibility}

\author{
 Sheng-Fu Wang \\
  Academia Sinica \\
  Institute of Linguistics \\
  Taipei, Taiwan \\
  \texttt{sftwang@gate.sinica.edu.tw} \\ \And
  Laurent Prévot \\
  CNRS \& MEAE \\
  CEFC\\
  Taipei, Taiwan\\
  \texttt{laurent.prevot@cnrs.fr} \\\AND
   Jou-An Chi \\
  Graduate Institute of Linguistics \\
  National Taiwan University \\
  Taipei, Taiwan \\
  \texttt{r11142005@ntu.edu.tw} \\\And
Ri-Sheng Huang \\
Department of CSIE \\
National Taiwan University \\
Taipei, Taiwan \\
 \texttt{r13922102@csie.ntu.edu.tw}   \\\And
  Shu-Kai Hsieh \\
  Graduate Institute of Linguistics \\
  National Taiwan University \\
  Taipei, Taiwan  \\
  \texttt{shukaihsieh@ntu.edu.tw} \\}

\begin{document}

\maketitle

\begin{abstract}
The achievements of {\it Large Language Models} in Natural Language Processing, especially for high-resource languages, call for a better understanding of their characteristics from a cognitive perspective.
Researchers have attempted to evaluate artificial models by testing their ability to predict behavioral (e.g., eye-tracking fixations) and physiological (e.g., brain responses) variables during language processing (e.g., reading/listening). In this paper, we propose using spontaneous speech corpora to derive production variables (speech reductions, prosodic prominences) and applying them in a similar fashion. More precisely, we extract.  We then test models trained with a standard procedure on different pretraining datasets (written, spoken, and mixed genres) for their ability to predict these two variables. Our results show that, after some fine-tuning, the models can predict these production variables well above baselines. We also observe that spoken genre training data provides more accurate predictions than written genres. These results contribute to the broader effort of using high-quality speech corpora as benchmarks for LLMs.
\end{abstract}

\section{Introduction}

The success of {\it Large Language Models} in Natural Language Processing, especially for high-resource languages, calls for a better understanding of their characteristics from a cognitive perspective. Yet, the difference in scale between the data size required to train LLMs and the amount of linguistic exposure needed for humans to achieve language proficiency calls for a better understanding of these models from a cognitive viewpoint. Researchers have attempted to evaluate artificial models by testing their ability to predict behavioral (e.g., eye-tracking fixations) and (neuro-)physiological (e.g., brain responses) variables during language processing (e.g., reading/listening). In this paper, we propose using spontaneous speech corpora to derive production variables and applying them in a similar fashion. More precisely, we extract {\it speech reduction} and {\it prosodic prominences} from three corpora in different languages (English, French, and Taiwan Mandarin).
Speech reductions shortens words duration by dropping sounds or merging syllables. Prosodic prominence makes certain syllables or words stand out through prosodic marking.

The {\sc BabyLM} initiative \cite{choshen2024call, hu2024findings, warstadt2023findings, warstadt2023call} has created an impulse for creating and evaluating cognitively plausible LLMs, both for the machine learning purpose of designing an efficient model with a small amount of training data (10-100M tokens) and for the cognitive science purpose of better understanding the similarities and differences between artificial and human language learning.  An observation about the initiative to date is that the datasets used are all in English. Although this is a natural starting point, it represents a significant limitation. Expanding the scope to include more languages is not only about better representing linguistic communities or potential model users; it is also about achieving comparable, contrastive results across different languages, which should offer valuable insights into both the learning models and the underlying learning processes.

While gathering a 100M token conversational dataset based on real spoken conversational data might be challenging, a 10M token dataset is accessible for languages like English, French, Mandarin, and some other languages studied from a corpus linguistics perspective. Conversational speech is the genre within which humans acquire their basic language skills. It is a genre quite distant from the usual written or web content on which LMs are trained, increasing the risk of biases in the LMs produced. Moreover, it has been argued that this genre is highly relevant to language emergence \cite{levinson2020human,Christiansen2022language}. How could a purely interactional dataset, including both child-directed and general conversation transcripts, be compared to written genres or more balanced mixtures \cite{feng2024child}?

In this context, current evaluation metrics, such as the ones used in {\sc babyLM}, while a good starting point, appear biased in two ways: they tend to favor canonical written forms and to prioritize syntactic, semantic, and commonsense pragmatics. However, language and communicative competence include many other dimensions. Although the initiative clearly emphasizes the importance of using speech transcripts—both child-directed and everyday conversations—as training data, to our knowledge, none of the evaluation metrics employed explicitly address the specificities of spontaneous speech.

To summarize, we argue that datasets and evaluation metrics are just as crucial as models for understanding the computational learning of language structure. More precisely, researchers have attempted to evaluate artificial models by testing their ability to predict behavioral (e.g., eye-tracking fixations) \citep{hollenstein2021cmcl} and physiological (e.g., brain responses) \citep{bingel2016extracting,hollenstein2018zuco,pasquiou2022neural} variables during language processing (e.g., reading/listening). We propose using spontaneous speech corpora to derive production variables and applying them in a similar fashion. More precisely, we extract {\it speech reduction'} and {\it prosodic prominences'} from three corpora in different languages (English, French, and Taiwan Mandarin). The three languages were chosen because of the availability of high-quality speech corpora (from which these variables can be reliably extracted), the availability of enough spontaneous speech transcripts (to train the model), and to ensure the inclusion of languages from different typological families.
We then test models trained with a standard procedure on different pretraining datasets (written, spoken, and mixed genres) to assess their ability to predict these two variables.
Our results show that the models can, after some fine-tuning, predict these production variables well above baselines. We also observe that spoken genre training data provides more accurate predictions than written genres. These results open the possibility of using high-quality speech corpora as benchmarks for LLMs.

\section{Related Work}
\label{sec:soa}
Since the emergence of large language models, there has been strong interest in the computational linguistics community in understanding why they are so successful. \citet{warstadt-etal-2020-learning} explored the conditions (e.g., the amount of training data) under which {\sc RoBERTa} develops and leverages linguistic features, such as part of speech (POS) and morphology, as opposed to relying on simpler surface-level features like position-based or length-based features.
More recently, several studies have probed LLMs to better characterize their performance across various domains, particularly with regard to their linguistic competence versus commonsense reasoning. These studies have also examined the relationship between model performance and the amount of training data required for different tasks.
In particular, \citet{zhang-etal-2021-need} used training sets of varying sizes—1M, 10M, 100M, and 1B tokens—to show that syntactic and semantic competence become robust within the 10M-100M range, whereas larger datasets are needed to achieve strong results in pragmatic and commonsense reasoning tasks.

More broadly, there have been proposals for evaluating the performance of LLMs on diverse linguistic tasks. \citet{warstadt-etal-2019-neural} leveraged a substantial body of generative syntax-semantics literature to develop benchmarks based on acceptability judgments, coming from the linguistic literature, like the {\sc CoLA} benchmark, which was further extended by exploiting more sources and data augmentation methods in {\sc BLiMP} \cite{warstadt2020blimp}.
In addition to these binary decision tasks, \citet{zhang-etal-2021-need} combined three other types of evaluation metrics: {\it classifier probing} (following \cite{ettinger2016probing,adi2017fine}), which includes tasks from POS tagging to coreference resolution; {\it information-theoretic} probing based on the minimum description length (MDL) principle; and {\it fine-tuning on higher-level tasks} such as those in the {\sc SuperGLUE} benchmark.

Most of the benchmarks have been proposed for English. However, {\sc BLiMP} \citet{Warstadt_2019} has inspired a series of language-specific benchmarks, such as {\sc Climp} for Mandarin Chinese \cite{xiang-etal-2021-climp}, as well as benchmarks for other languages like Japanese \cite{someya2023jblimp}, Dutch \cite{suijkerbuijkblimp}, and Russian \cite{taktasheva2024rublimp}. These are important additions to the evaluation landscape.
While these benchmarks represent important extensions to the general evaluation framework, they all rely on syntax-semantics structures derived from introspection and textbook data, as will be discussed in the next section. In parallel with these efforts, monolingual language models have been developed using large amounts of data \cite{chang2024goldfish}, as well as experiments involving varied data quantities \cite{micheli-etal-2020-importance}.

In another line of research, several studies have tested the ability of large language models (LLMs) to perform tasks inspired by cognitive science, particularly in the domains of semantics and pragmatics \cite{Ettinger_2020,binz2023using}.

Our approach of using actual speech data to extract production-based metrics can be related to studies that use behavioral or neurophysiological data linked with linguistic datasets. Specifically, there has been significant work focusing on textual datasets combined with eye-tracking  \cite{hollenstein2021cmcl} or neurophysiological \cite{bingel2016extracting,hollenstein2018zuco} measures. Additionally, datasets from passive listening tasks, linked to fMRI, have been released for various languages (e.g., French, Mandarin, and English) \cite{li2022petit}. These datasets have been used, for instance, to study the impact of training parameters on a language model's ability to predict neurophysiological data \cite{pasquiou2022neural}. Focusing on spontaneous speech, \cite{rauchbauer2019brain,hmamouche2024interpretable} examined the predictability of fMRI-derived signals from conversational variables, including lexical information.

In terms of specialized language models, \cite{cabiddu2025comparing} developed LMs based on child-directed speech transcripts and evaluated them on word-sense disambiguation tasks. They concluded that word acquisition trajectories could be better captured by multimodal models that incorporate acoustic features, among other aspects. Regarding tokenizers, \citet{beinborn2023analyzing} proposed an evaluation paradigm focusing on the cognitive plausibility of subword tokenization. They compared {\tt BPE}, {\tt WordPiece}, and {\tt UnigramLM} and revealed a lower ``cognitive correlation'' for the latter. Furthermore, in the most recent {\sc BabyLM} edition, \cite{martinez2023climb} introduced an interesting learning curriculum that constrained vocabulary in the early stages to simulate more cognitively plausible learning curves. Although this approach did not yield consistent overall results, marginal gains were observed in selected tasks. 

Lastly, a recent audio-based question-answering benchmark for spoken language models incorporates spontaneous speech phenomena such as filler words, disfluencies, false starts, and corrections into different ``speech style'' conditions for testing the robustness of these models \cite{cui2025voxeval}. Notably, the injection of speech phenomena into the benchmark is text-based and LLM-generated instead of being collected from natural speech.

\section{A proposal for a new source of metrics}
\label{sec:metrics}

Most of the work mentioned above is grounded in text-based and/or handcrafted paradigms, potentially coupled with behavioral or physiological lab measures. In contrast, we propose using actual spontaneous conversational transcripts to build complementary benchmarks that test not only syntactic-semantic dimensions but also real-world language use. These metrics will remain fundamentally linguistic in nature rather than focusing on task-specific or end-to-end evaluation.

Language is acquired, especially in its early stages, within spontaneous, conversational environments. While conversational language shares grammatical structures with other genres, its unique characteristics suggest that simply listing syntactic ``errors'' or semantic incongruities does not fully capture linguistic competence. Moreover, in a conversational context, what may be considered a production error from a formal grammatical perspective is often perfectly acceptable and successfully achieves its communicative purpose. Therefore, we aim to develop a complementary approach that provides a broader set of metrics to evaluate language models from both cognitive and communicative perspectives along with existing benchmarks.

Specifically, we propose using spontaneous speech corpora, as they offer insights into human language processing through various observable production phenomena. Our approach is a kind of {\it classifier probing} \cite{ettinger2016probing,adi2017fine,warstadt-etal-2019-neural}, but rather than focusing on metalinguistic tasks (e.g., predicting syntactic categories), we aim to predict phenomena that serve as partial indicators of language processing. We propose two preliminary metrics:
{\it speech reductions} and {\it prosodic prominences}, which are grounded in spontaneous speech production, and each have been the subject of extensive research.

\subsection{Speech reductions}
\label{speech}
Speech reductions have been studied across a range of linguistic levels, especially when considering the issue of signal information density. In spontaneous speech, some chunks of speech are produced in a reduced manner, both in terms of duration and articulatory amplitude. The location of these reductions is not random. For example, studies have suggested that speakers tend to smooth the information density of their speech signal over time, with reductions serving as a mechanism to achieve this smoothing effect \cite{aylett2004smooth}.

The relationship between information density and speech reduction has led to research developments on this topic with various approaches. These approaches may differ in the probabilistic measures used to predict reductions, such as lexical frequency, contextual probability, and informativity \cite{aylett2004smooth, gahl2008time, priva2012sign, seyfarth2014word}. They also differ in terms of the linguistic level at which reductions occur, whether at the phoneme-, syllable-, word-level, or in terms of overall speech rate. Many of these studies include and compare different types of probabilistic measurements (e.g., lexical frequency and contextual probability) within a single study (e.g., \cite{seyfarth2014word, priva2018interdependence}) and some of them also compare probabilistic measurements calculated at different linguistic levels (e.g., segment- and syllable-levels in \citet{van1999effects}, segment- and word-level measurements in \citet{van2003efficient}, syllable- and word-level measurements in \citet{wang2022interaction}). Inclusion and comparison of reductions or phonetic variability across various linguistic levels in the same study have also been done (e.g., for individual segments and prefixes as a whole in \citet{pluymaekers2005lexical}; morphemes and words in \citet{tang2018contextual}), albeit less frequently.

These studies show that phonetic reduction can be predicted to varying degrees based on the statistical distribution of linguistic units, and the prediction has been repeatedly found with different types of measurements at various levels of linguistic units. This motivates the development of a reduction-labeling task for evaluating language models.

\subsection{Prosodic Prominences}

Prosodic prominence refers to the emphasis placed on certain units, often demarcated at the level of words or syllables, within a spoken utterance. This emphasis can be measured through (and perceived based on) acoustic cues such as movements in fundamental frequencies, duration, intensity, and segmental properties such as the formant structure of vowels. Recent work by \citet{wolf-etal-2023-quantifying} has shown a significant degree of redundancy between the representations encoded from tokens alone and those derived from acoustic-prosodic information. Acoustic-prosodic features such as word-level energy, fundamental frequency, duration, pause, and composite measurements derived using a wavelet-based algorithm \cite{suni2017hierarchical} were used to quantify this redundancy. Their findings suggest that prosodic information can be predicted, to some extent, from the word itself and its surrounding context. There are also studies showing that language models (e.g., {\sc BERT}) can predict prosodically-prominent tokens based from speech transcripts \cite{kakouros2023does, talman2019predicting}. These findings suggest that part of prosodic prominence can indeed be predicted by a language model, even though predicting based on text alone should never be sufficient (as also noted by \citet{wolf-etal-2023-quantifying}).

\section{Method}
\label{sec:data}

\subsection{Pre-training LLMs}

We pretrained multiple RoBERTa models (an architecture choice motivated in part by a more direct comparison of our topline choice, XLM-RoBERTa, which has a more manageable size for fine-tuning) using three corpora in different languages: English, French, and Taiwan Mandarin. For each language, the pretraining data is categorized into three distinct types: conversational, written, and a mix of these two genres. For English, the spoken data was drawn from BNC and Switchboard, and the written data was drawn from Simple Wikipedia (both extracted from the {\sc BabyLM} 100M dataset). The mixed data is a 9M-token subset of the {\sc BabyLM} 10M training data, which contains CHILDES and Gutenberg collections in addition to the three aforementioned sources. The 1M-token dev sets were built from the corresponding validation sets in the {\sc BabyLM} data. 

For French, the conversational data consists of 10 million tokens from conversational datasets specifically {\sc ORFEO} \cite{benzitoun2016projet} and {\sc CHILDES-fr} \cite{macwhinney2014childes,rose2014phonbank}, and the conversational data contains an additional 10 million tokens sourced from Wikipedia. 

For Taiwan Mandarin, the 10M-token conversational data was derived from the NCCU Spoken Corpus of Taiwan Mandarin \cite{chui2008nccu}, Taiwan Corpus of Child Mandarin (TCCM)\footnote{\url{https://lope.linguistics.ntu.edu.tw/tccm/}} \cite{chang2011tccm}, Open Subtitles \cite{lison2016opensubtitles2016}, and the transcript of Taiwan Legislative Yuan \footnote{\url{https://www.parliamentarytv.org.tw/}}. The written data was from a subset of Traditional Chinese Wikipedia \footnote{\url{https://huggingface.co/datasets/zetavg/zh-tw-wikipedia/}}. 
For French and Mandarin, the mixed data was built from a 50-50 mix of the conversational and Wiki data, and the prepared datasets for all three types underwent a 90-10 split to build the training and development sets. 

We employed the {\it SentencePiece} tokenizer \cite{kudo2018sentencepiece} with the Unigram model \cite{kudo2018subword}. SentencePiece allows tokenization without requiring predefined word boundaries, making it particularly suitable for languages with non-segmented scripts, such as Taiwan Mandarin. Additionally, we applied a minimum token frequency threshold of 2. The vocabulary size was set to 10,000 tokens, which was determined empirically based on preliminary experiments. Each model was initialized with XLM-RoBERTa as the pretrained base model and trained for 100 epochs using a batch size of 32. The English, French, and Mandarin models were pretrained with a learning rate of 1e-4, 1e-4, and 2e-4 respectively, with early stopping if validation loss failed to decrease in 10 epochs.

\subsection{Benchmarks}
\label{sec:benchmarks}

For these experiments, we used three sources to build benchmarks: the Buckeye Corpus for English\footnote{\url{https://buckeyecorpus.osu.edu/}} \cite{pitt2005buckeye}, the Corpus of Interactional Data (CID) for French\footnote{\url{https://hdl.handle.net/11403/sldr000720}} \cite{blache2017corpus}, and The Sinica Mandarin Conversational Dialogue Corpus (Sinica MCDC8)\footnote{\url{https://www.aclclp.org.tw/use_mat.php\#mcdc}} for Mandarin \cite{tseng2013lexical}. CID is an 8-hour corpus of 1-hour conversations between friends (16 speakers). It features fiercely spontaneous conversational speech. Buckeye is a corpus with 38.1 hours of spontaneous speech (40 speakers) recorded in an interview format. MCDC8 features 8 one-hour natural conversations between recruited participants (16 speakers) who did not know their interlocutor in advance. The main reason for the choice of these corpora is the high quality of their speech transcript alignment, down to the syllable or even the segment level.

\subsubsection{Speech Reduction}

There are several methods to determine whether a portion of speech is reduced. Following approaches in the literature, we first derived ratios of every word token's actual duration and its expected duration. For the French and Mandarin benchmarks, we used annotations of syllable boundaries in the corpus and developed a model that predicts syllable duration based on the segment it contains, similar to \citet{wang2022interaction}. A model is trained on one-half of the corpus and then applied to estimate the expected token duration in the remaining half of the corpus. For the English benchmark, we followed the literature  \citep{bell2009predictability, gahl2012reduce, seyfarth2014word} and made use of the available segmental duration in the corpus and calculated a word' expected duration. The corpus was similarly divided in halves with expected duration in one half calculated with segment duration from the other.

In both cases, we then converted the ratios into binary labels by applying a threshold of 0.5 (i.e., a reduction of at least 50\%) for English and French, and a threshold of 0.6 (reduction of at least 40\%) for Mandarin. The threshold was adjusted for Mandarin to keep the proportion of reduced tokens more similar to other two languages, as Mandarin's syllable-timed nature make substantial reduction a rare event. These thresholds resulted in 13.58\%, 17.54\%, and 11.09\% of the tokens in English, French, and Mandarin being labeled as reduced. These labels were then encoded in BIO format.

\subsubsection{Prosodic Prominences}
\label{sec:prom}
To detect prosodically prominent tokens we used \citeauthor{suni2017hierarchical}'s (\citeyear{suni2017hierarchical}) method based on wavelet method that combines acoustic features (fundamental frequency, energy, duration) fr determining prominence at the token level. One of the reasons for this tool choice is that it has already been used in the LM literature \cite{wolf-etal-2023-quantifying} to quantify the amount of redundancy between textual and prosodic levels. We applied the default configuration of this tool and set a threshold score of $1.25$ (See figure \ref{fig:distrib-prom-fr} in Appendix \ref{appendix:distribution} for details of the score distribution), which yielded 14\%, 13.79\%, and 12.77\% of prosodic prominent tokens in English, French, and Mandarin.

\subsection{Fine-tuning Experiments}
\label{sec:expes}
The experiments evaluated different pre-trained models for our set of tasks.\footnote{Notebooks for pretraining LMs and performing the experiment can be accessed at ***.} 
More precisely, we fine-tuned the pretrained models separately on a token classification task to predict which tokens were labeled (reduced / prominent) and which were not. A simple cross-validation was conducted across groups of speakers to maximize diversity across the folds. The models were fine-tuned for 10 epochs, with the validation fold being used for early stopping (patience = 5). For each model-task combination, the fine-tuning experiment was run for 10 iterations, i.e.,  five learning rates (2e-5, 4e-5, 6e-5, 8e-5, 1e-4) $\times$ two batch sizes (32, 16), with the combination that resulted in the highest averaged f1 score across 8 folds being used in the analysis.

Finally, we include two conditions using fine-tuning {\sc XLM-RoBERTa} {\sc Base} and {\sc Large}. This condition is not fair with the other models, since it is pretrained on much larger data set. It was originally included both as a topline and to assess the feasability of the task. For more efficient fine-tuning, these models underwent vocabulary pruning \cite{yang-etal-2022-textpruner} on the training texts for our three languages.

\begin{figure*}[hbt]
    \centering
    \includegraphics[width=\linewidth]{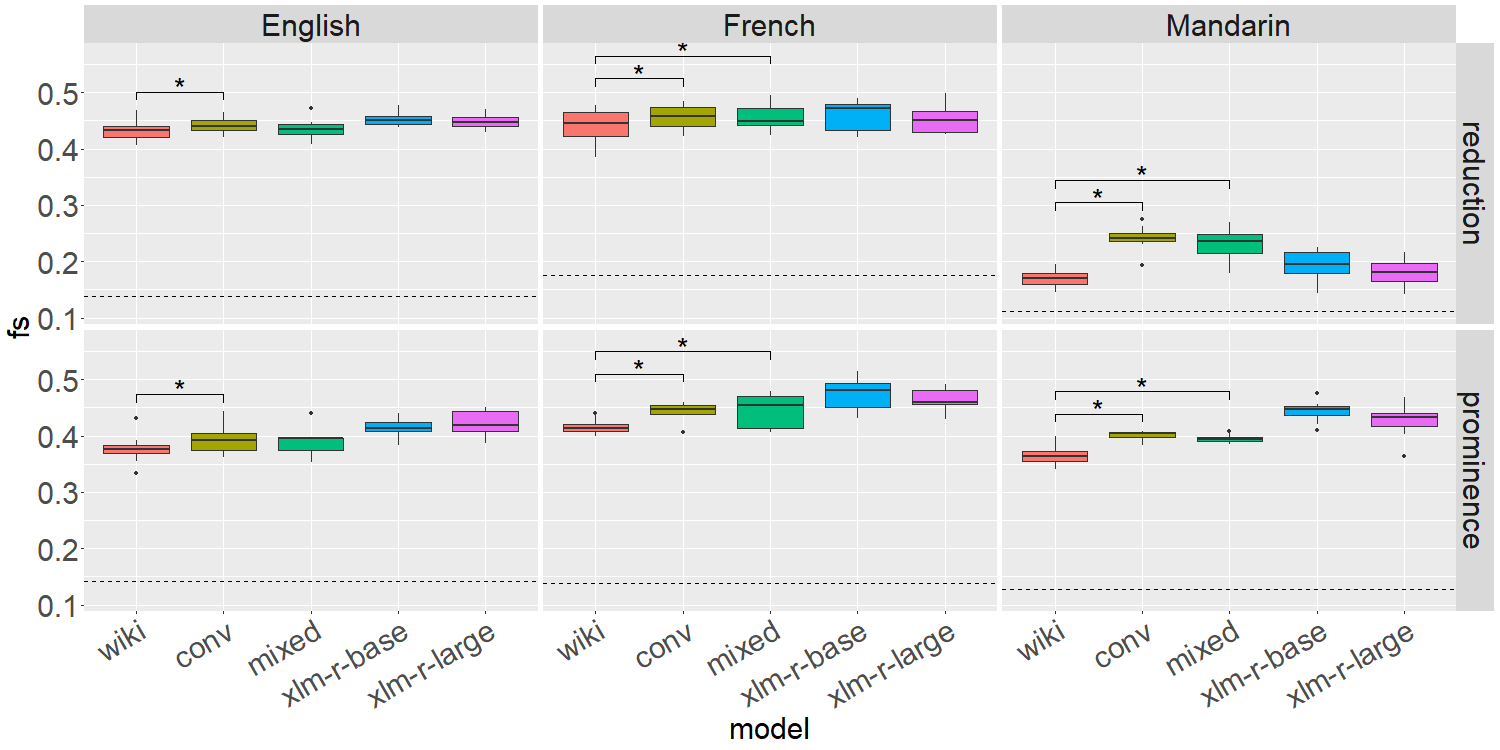}
    \caption{F-scores on the benchmarks as a function of model (x), task and language (prominence). Model comparisons are based on Bayesian regression analyses where MODEL is the fixed predictor and FOLD is a random intercept. The models were run with weak (uniform) priors using the \textit{brms} package in R, with post hoc hypothesis testing focused on comparing three small models (Wiki, conversational, mixed). Stars in the figure indicate a one-sided hypothesis with a posterior probability above 95\%. Dotted lines correspond to random baselines.}
    \label{fig:fscore_all}
\end{figure*}

\begin{figure}[!hbt]
    \centering
    \includegraphics[width=\linewidth]{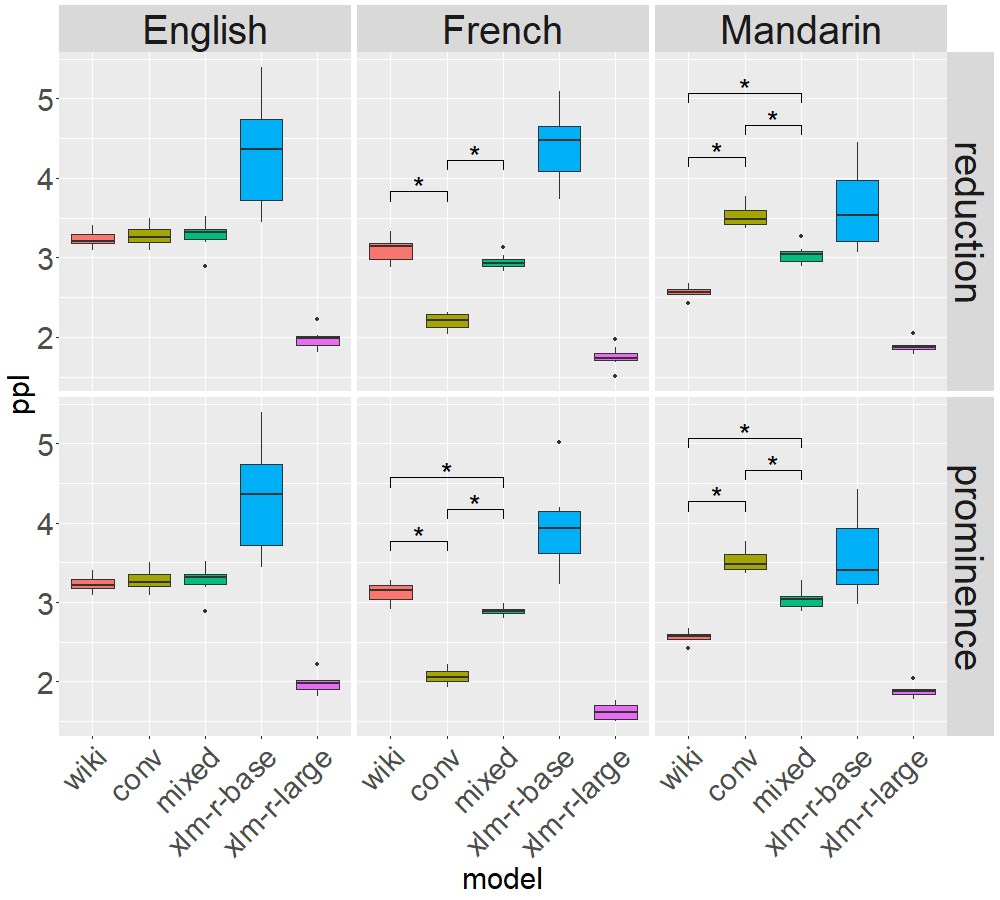}
    \caption{Perplexity on the benchmarks}
    \label{fig:ppl}
\end{figure}

\begin{figure}[!hbt]
    \centering
    \includegraphics[width=\linewidth]{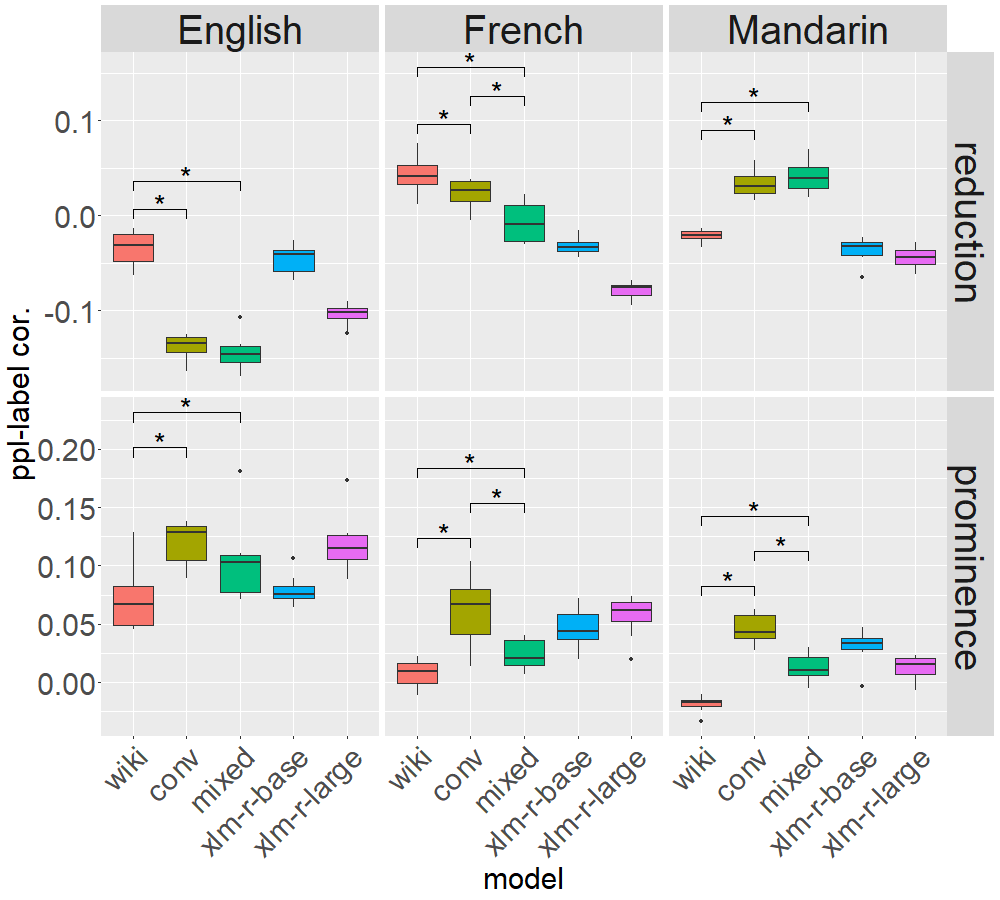}
    \caption{The correlation between perplexity and labels.}
    \label{fig:ppl_gold}
\end{figure}

\section{Results}

The main results are presented in Figure \ref{fig:fscore_all}. (see more detailed results in Appendix \ref{appendix:complete}). For reduction, the results indicate that models comfortably outperform the label distribution baselines (dotted lines), especially for English and French. Between models, in all three languages, the Wikipedia model is outperformed by at least the conversational model. Furthermore, the topline RoBERTa models do not exhibit obvious advantages over the smaller model. It is also worth noting that models perform much worse in Mandarin than in the other two languages.

As for prominence, the performance is more comparable across the three languages. There is also a similar advantage for conversational and mixed models over the Wikipedia model. Contrary to the reduction task, the topline {\sc RoBERTa} models exhibit an advantage over the smaller models. 

\subsection{Perplexity and benchmarking}
As language models' capacity is often evaluated on their perplexity on a given corpus, we calculated per-token perplexity (handling multi-token words with the masking method in \cite{kauf2023better}) for each language-model combination on each fold in the benchmarks, shown in Figure \ref{fig:ppl}.  For the smaller models, three languages exhibit distinct patterns when it comes to the difference between models: For English, the perplexity scores were very similar, while for French, the conversational model outperforms the other models in having the lowest perplexity, followed by the mixed model and the Wiki model. For Mandarin, the pattern is reversed with the Wiki model having the lowest perplexity.  In other words, for English and especially Mandarin, there is a divergence of perplexity scores between the benchmark and the fine-tuning experiment.

We also examined the correlation between token-level surprisal and gold labels, an analysis similar to the aforementioned studies that examine whether surprisal/perplexity predicts behavioral measurements such as reading time. The results are shown in Figure \ref{fig:ppl_gold}. Based on observations from phonetic research, perplexity's correlation is expected to be negative for both reduction and prominence \cite{aylett2004smooth, seyfarth2014word}. For prominence, the correlation is more positive for conversational and mixed models as observed in the fine-tuning experiments, thereby suggesting that models' fine-tuning performance on categorical prominence prediction is partially influenced by the behavior of the base models.

For reduction, the connection is less established: The expected negative correlation between perplexity and reduction is only borne out for conversational and mixed models in English, even though the relative difference between Wiki and conversational/mixed models in French is similar. Notably, the Wiki model is the only small model that showed the expected correlation in Mandarin. As these patterns are not perfect matches of the fine-tuning results, it suggests that at least for reduction, the fine-tuning task is not as trivial as something that can be predicted from base models' behavior.

These observations, summarized in terms of the winner of the comparison between the conversational and Wiki models in Table \ref{tab:winners}, paint a more nuanced picture of the relationship between fine-tuning performance, perplexity, and how perplexity predicts reduction and prominence. They also add to the previous discussions on how and whether better language models in perplexity do not guarantee better cognitive modeling \cite{kuribayashi2021lower}.

\begin{table}[hbt]
\centering
\caption{Winners of the comparison between the conversational and Wiki models in different criteria: higher F1 in fine-tuning (FT); lower perplexity (ppl), lower perplexity-label correlation (ppl-label cor.) for reduction and higher correlation for prominence.}
\label{tab:winners}

\begin{tabular}{|l|l|l|l|l|}
\hline
lge       & task  & FT & ppl & ppl-label cor. \\ \hline
Eng.  & both  & conv        & n.s. & conv     \\ \hline
Fre.   & both  & conv        & conv & conv     \\ \hline
Man. & reduc & conv        & wiki & wiki    \\ \hline
Man.  & prom  & conv        & wiki & conv      \\ \hline       
\end{tabular}
\end{table}

\subsection{Error Analysis}


To better understand model errors in predicting prominence and reduction, we calculated the most over-predicted and under-predicted words for each language based on the difference between the models' predictions and gold label rates, as shown in Appendix \ref{appendix:word}. The variances are very low across models, showing that models with different pre-training data have similar tendencies in over- and under-predictions.

Across languages, the most over-predicted words for reduction are function words and discourse markers. As for prominence, over-predicted words include discourse markers and content words directly linked to the topic of the discourse. Under-predicted reductions are mostly content words and less commonly reduced expressions For prominence, under-predicted words also tend to be content words. We refer the readers to Appendix \ref{appendix:word} for examples of these words.




A general explanation may be that models have learned that some words are intrinsically more likely to be reduced \cite{kubra2023reduction} or to be prominent, and therefore biased towards the more common class. The tendency is likely related to lexical frequency and distribution of words within the discourse structure. It is worth noting that even though over- and under-predictions are observed, the overall correlation between frequency and label assignments did not differ greatly between model predictions and gold labels (Figures \ref{fig:freq_reduc} and \ref{fig:freq_prom} in the Appendix \ref{appendix:freq_labels}).


\section{Potential shortcomings and Limitations}
\label{sec:limits}

{\bf Information-centric nature.} $\quad$
One potential limitation is that the models may only capture the information-theoretic contribution to our tasks, i.e., the well-noted relationship between information-theoretic notions such as information density, entropy, and predictability and phonetic realizations. However, the prediction of these phenomena cannot be reduced to information-theoretic explanations alone, and our analyses do show some divergences between fine-tuning experiments and direct correlations between perplexity and benchmark labels. As each metric introduces its own set of subtleties related to language processing, our goal is to evaluate LLMs in terms of their ability to grasp these subtleties.

{\bf Text-only.} $\quad$ The proposed phenomena for probing the models are inherently related to speech processing, which goes beyond what can be achieved with a text-only approach. Beyond the acoustic modality, the visual channel also plays a role. Our goal in proposing these metrics is not to achieve state-of-the-art performance in predicting these phenomena. Rather, we aim to treat them as {\it `traces'} of human language processing visible at the surface level, and to test which models are better at predicting these traces from text-only input.

{\bf Surface level shortcuts.}  $\quad$ 
A concern related to the previous point is the risk that models may rely on surface-level elements as shortcuts to predict the target variables. Addressing such a concern is inherently challenging, as speech phenomena are deeply intertwined with observable surface patterns. Nonetheless, we believe it is still worth pursuing this line of investigation, which can be extended particularly through controlled evaluation sets, similar to the approach in \citet{mccoy-etal-2019-right}, who compared the models with some heuristics to isolate the influence of surface-level cues. The types of cross-lingual phenomena in our proposed benchmarks also invite future work that develops principled methodologies in evaluating the `cognitive' process inside a language model.

{\bf Triviality of the main result.} From a machine-learning perspective, it might be seen as a trivial result that models trained on data similar to test sets perform better than models trained on other types of data. First of all, it is worth emphasizing that pretraining datasets and benchmarks in our experiments are completely independent as they do not come from the same raw corpora. Also, the pretraining datasets and corpora for building benchmarks have been curated by different teams and transcribed following different conventions. Nevertheless, we cannot deny that the conversational datasets are by all aspects (sentence length distribution, lexical frequencies, etc) more similar to benchmarks than Wikipedia datasets are.

As trivial as it seems, it may be one of our main points: to produce models more closely related to human cognition, one should use data sets made of spontaneous speech (and not generic textual / web content). Furthermore, the additional analyses on perplexity and how perplexity directly correlates with labels in benchmarks show that the advantage of conversational models in these conversational benchmarks is not guaranteed in all potential ways of `evaluating' these models.

\section{Conclusion}

In this paper, we propose advancing the benchmarking of language models to include spontaneous speech phenomena and to extend beyond English. Motivated by the {\sc BabyLM} initiative, we evaluated models trained on developmentally plausible amounts of data and showed how models trained in different genres of texts perform differently. 

In the future, it will be crucial to explore more nuanced variations in training data, such as balancing conversational speech, child-directed speech, and simple texts as well as developping complementary evaluation metrics. 

From a broader perspective, we hope to show that benchmarks like {\sc BLiMP} that require a significant amount of expert and naive human input to build, can be complemented, as we have explored in this paper, with benchmarks derived from existing high-quality linguistic corpora, without additional human efforts.

\section*{Acknowledgments}
We would like to thank Shu-Chuan Tseng and Pierre Magistry for discussions in relation to this paper. This study was supported by Taiwan's National Science and Technology Council (NSTC-112-2410-H-001-098-MY2).

\bibliography{babylm}

\clearpage

\appendix

\section{Distribution of benchmark labels}
\label{appendix:distribution}

\begin{figure}[!htb]
    \centering
    \includegraphics[width=\linewidth]{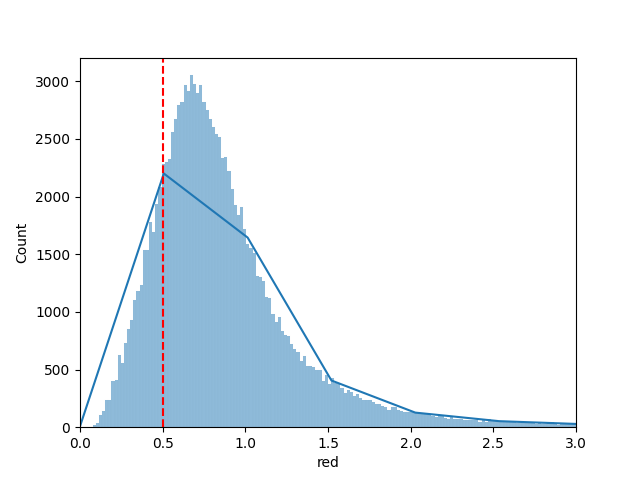}
    \caption{Distribution reduction ratios as calculated in the English Dataset and the threshold selected.}
    \label{fig:distrib-reduc-en}
\end{figure}

\begin{figure}[!htb]
    \centering
    \includegraphics[width=\linewidth]{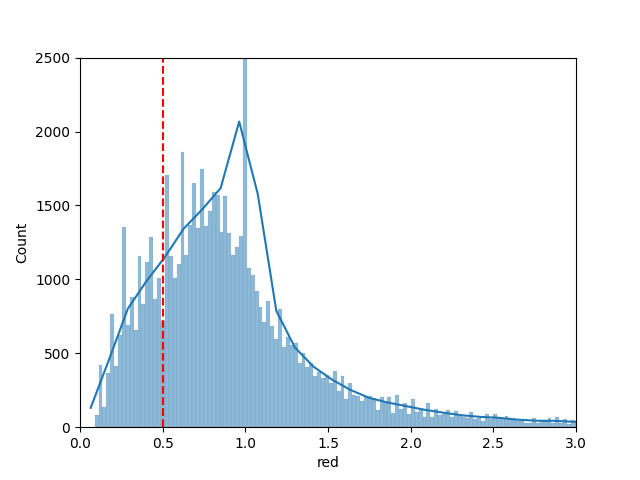}
    \caption{Distribution reduction ratios as calculated in the French Dataset and the threshold selected.}
    \label{fig:distrib-reduc-fr}
\end{figure}

\begin{figure}[!htb]
    \centering
    \includegraphics[width=\linewidth]{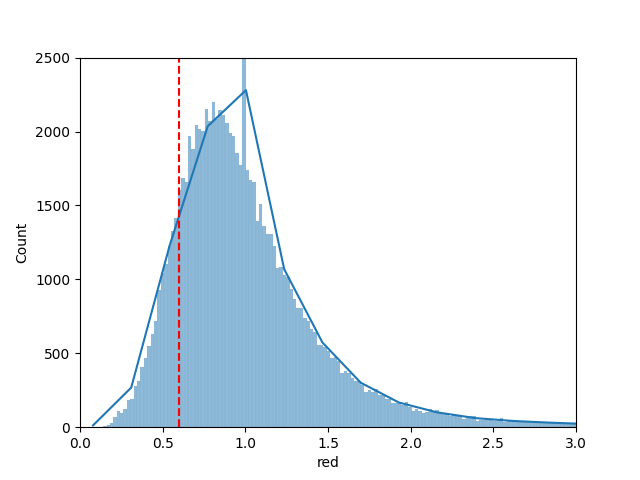}
    \caption{Distribution of reduction ratios as calculated in the Mandarin Dataset and the threshold selected.}
    \label{fig:distrib_reduc-zh}
\end{figure}

\begin{figure}[!htb]
    \centering
    \includegraphics[width=\linewidth]{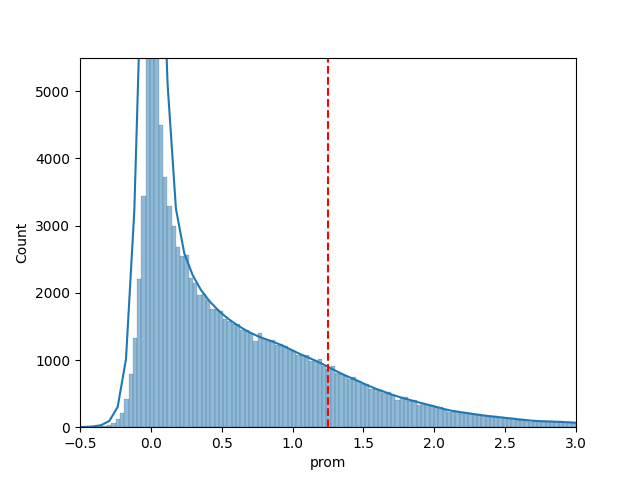}
    \caption{Distribution of prominence scores as calculated in the English Dataset and the threshold selected.}
    \label{fig:distrib-prom-en}
\end{figure}

\begin{figure}[!htb]
    \centering
    \includegraphics[width=\linewidth]{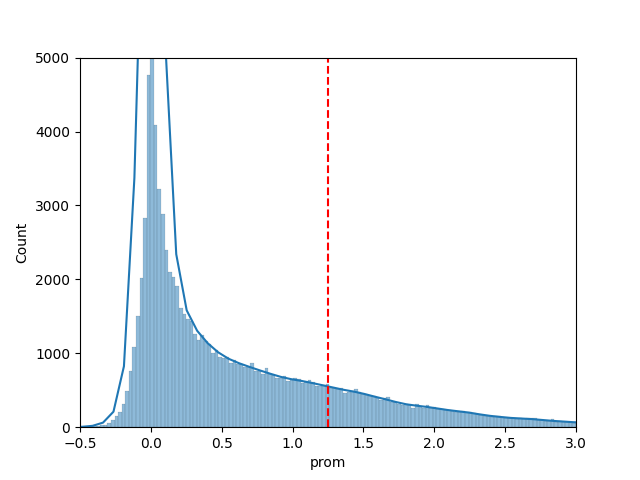}
    \caption{Distribution of prominence scores as calculated in the French Dataset and the threshold selected.}
    \label{fig:distrib-prom-fr}
\end{figure}

\begin{figure}[!htb]
    \centering
    \includegraphics[width=\linewidth]{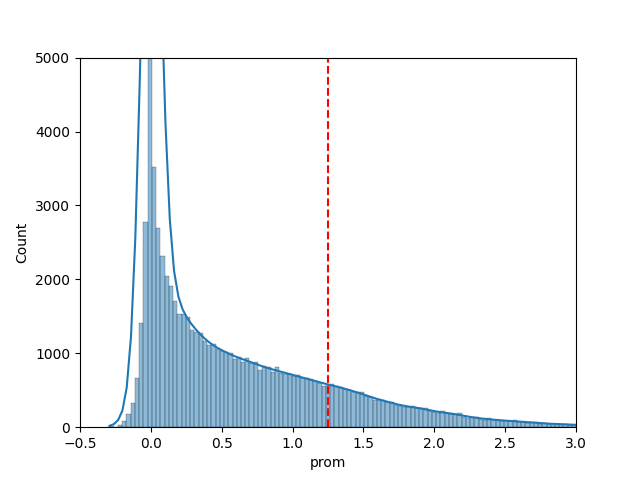}
    \caption{Distribution of prominence scores as calculated in the Mandarin Dataset and the threshold selected.}
    \label{fig:distrib-prom-zh}
\end{figure}

\newpage
\onecolumn

\section{Complete scores}

\label{appendix:complete}

\begin{table*}[!htb]
  \centering
    \caption{Full results on the proposed speech-based benchmarks}
  \label{tab:1}
  \begin{tabular}{|l|l|l|l|l|l|} 
  \hline
 Language & Task & Model &   F1  & Precision  &  Recall   \\  \hline
English & reduction & wiki & .433 (.020) & .451 (.038) & .422 (.052) \\ \cline{3-6} 
 &  & conv & .442 (.014) & .465 (.023) & .426 (.041) \\ \cline{3-6} 
 &  & mixed & .436 (.019) & .448 (.025) & .429 (.048) \\ \cline{3-6} 
 &  & xlm-r-base & .452 (.013) & .458 (.029) & .451 (.041) \\ \cline{3-6} 
 &  & xlm-r-large & .448 (.013) & .433 (.033) & .473 (.055) \\ \cline{2-6} 
 & prominence & wiki & .378 (.028) & .407 (.035) & .356 (.044) \\ \cline{3-6} 
 &  & conv & .394 (.027) & .428 (.027) & .366 (.041) \\ \cline{3-6} 
 &  & mixed & .390 (.025) & .414 (.040) & .372 (.035) \\ \cline{3-6} 
 &  & xlm-r-base & .414 (.018) & .463 (.046) & .378 (.032) \\ \cline{3-6} 
 &  & xlm-r-large & .422 (.024) & .447 (.039) & .405 (.051) \\ \cline{1-6} 
French & reduction & wiki & .441 (.034) & .472 (.051) & .415 (.031) \\ \cline{3-6} 
 &  & conv & .455 (.023) & .488 (.036) & .429 (.038) \\ \cline{3-6} 
 &  & mixed & .456 (.024) & .498 (.038) & .424 (.035) \\ \cline{3-6} 
 &  & xlm-r-base & .460 (.027) & .491 (.047) & .437 (.042) \\ \cline{3-6} 
 &  & xlm-r-large & .454 (.028) & .473 (.036) & .438 (.030) \\ \cline{2-6} 
 & prominence & wiki & .416 (.015) & .415 (.039) & .422 (.032) \\ \cline{3-6} 
 &  & conv & .443 (.017) & .453 (.035) & .436 (.032) \\ \cline{3-6} 
 &  & mixed & .445 (.031) & .456 (.030) & .435 (.045) \\ \cline{3-6} 
 &  & xlm-r-base & .474 (.030) & .478 (.037) & .472 (.038) \\ \cline{3-6} 
 &  & xlm-r-large & .464 (.020) & .463 (.036) & .468 (.029) \\ \cline{1-6} 
Mandarin & reduction & wiki & .169 (.016) & .154 (.023) & .195 (.038) \\ \cline{3-6} 
 &  & conv & .241 (.024) & .248 (.044) & .239 (.033) \\ \cline{3-6} 
 &  & mixed & .230 (.029) & .251 (.051) & .220 (.042) \\ \cline{3-6} 
 &  & xlm-r-base & .192 (.031) & .179 (.036) & .215 (.045) \\ \cline{3-6} 
 &  & xlm-r-large & .180 (.026) & .174 (.034) & .194 (.043) \\ \cline{2-6} 
 & prominence & wiki & .365 (.019) & .345 (.023) & .390 (.033) \\ \cline{3-6} 
 &  & conv & .401 (.008) & .398 (.018) & .405 (.025) \\ \cline{3-6} 
 &  & mixed & .395 (.008) & .379 (.019) & .413 (.015) \\ \cline{3-6} 
 &  & xlm-r-base & .444 (.020) & .447 (.029) & .444 (.037) \\ \cline{3-6} 
 &  & xlm-r-large & .426 (.031) & .429 (.037) & .425 (.036) \\ \cline{1-6} 
 
 \end{tabular}

\end{table*}

\newpage

\section{Word-level error analysis}
\label{appendix:word}
For each language and task, we measured the most over-predicted and the most under-predicted words by rate difference, which is the difference between the model prediction's true rate and the ground truth's true rate. Top five and bottom five words ranked by rate difference are listed below.
Since a single word may be split into multiple tokens, we determine a word’s prediction based on majority voting among its tokens. In the case of a tie, the word is classified as ``true'' (reduced/prominent). We only include words that appear at least 50 times.

\begin{table*}[htb]
  \centering
    \caption{The most over-predicted and under-predicted words for each task. For each pretraining scheme (wiki, conversational, mixed, {\sc RoBERTa}-base and {\sc RoBERTa}-large), we selected the best-performing model, and calculated every word's rate difference. Only the top-5 and bottom-5 words in rate difference are listed, along with the mean across the 5 models (and the standard deviation in parentheses).}

  \label{table:words}

  \begin{threeparttable}
  \begin{tabular}{|l|l|l|l|l|l|} 
  \hline

Language & Task & Top-5 & Rate difference & Bottom-5 & Rate difference \\  \hline
English & reduction & youre & .284 (.021) & u\tnote{1} & -.255 (.033) \\ \cline{3-6} 
 &  & dont & .219 (.033) & everybody & -.232 (.070) \\ \cline{3-6} 
 &  & theyre & .209 (.063) & only & -.232 (.111) \\ \cline{3-6} 
 &  & a & .179 (.008) & ill & -.226 (.057) \\ \cline{3-6} 
 &  & at & .170 (.022) & many & -.209 (.055) \\ \cline{2-6} 
 & prominence & basically & .215 (.082) & vote & -.225 (.023) \\ \cline{3-6} 
 &  & obviously & .165 (.145) & grade & -.144 (.047) \\ \cline{3-6} 
 &  & okay & .134 (.022) & wanted & -.126 (.036) \\ \cline{3-6} 
 &  & yeah & .121 (.048) & ten & -.125 (.044) \\ \cline{3-6}  
 &  & um & .110 (.027) & times & -.125 (.024) \\ \cline{1-6} 
 
French & reduction & jesuis & .302 (.036) & ont & -.183 (.047) \\ \cline{3-6} 
 &  & jesais & .236 (.012) & vais & -.180 (.071) \\ \cline{3-6} 
 &  & peutêtre & .205 (.026) & vois & -.160 (.022) \\ \cline{3-6} 
 &  & ai & .160 (.041) & oh & -.154 (.021) \\ \cline{3-6} 
 &  & de & .147 (.031) & te & -.149 (.036) \\ \cline{2-6} 

 & prominence & mhm & .352 (.179) & etc & -.167 (.108) \\ \cline{3-6}  
 &  & problème & .185 (.074) & école & -.144 (.107) \\ \cline{3-6} 
 &  & accord & .178 (.055)  & soit & -.117 (.048) \\ \cline{3-6}  
 &  & gamin & .161 (.051) & oh & -.115 (.069) \\ \cline{3-6}  
 &  & moment & .134 (.039) & pense & -.106 (.037) \\ \cline{1-6}

Mandarin & reduction & de0 & .201 (.191) & xue2-xiao4 & -.241 (.094) \\ \cline{3-6} 
 &  & ke3-shi4 & .192 (.059) & zhe4-ge0 & -.089 (.015) \\ \cline{3-6} 
 &  & yin1-wei1 & .148 (.044) & bian4-cheng2 & -.084 (.040) \\ \cline{3-6} 
 &  & le0 & .143 (.072) & da4-jia1 & -.079 (.040) \\ \cline{3-6} 
 &  & suo3-yi3 & .142 (.053) & bu4-neng2 & -.070 (.048) \\ \cline{2-6} 
 & prominence & dang1-bing1 & .272 (.044) & san1 & -.136 (.049) \\ \cline{3-6} 
 &  & xue2-xiao4 & .183 (.104) & jiao4 & -.131 (.027) \\ \cline{3-6} 
 &  & wei4-shen2-me0 & .119 (.057) & na4-me0 & -.091 (.047) \\ \cline{3-6} 
 &  & ru2-guo3 & .112 (.102) & dong1-xi1 & -.079 (.019) \\ \cline{3-6} 
 &  & da4-jia & .105 (.049) & che1 & -.076 (.011) \\ \cline{1-6}



 \end{tabular}
  \begin{tablenotes}
\item[1] u as in the initialism ``OSU (Ohio State University)'', which appears frequently in the Buckeye corpus.
\end{tablenotes}
\end{threeparttable}
\end{table*}

\section{Token frequency vs. gold and predicted labels}
\label{appendix:freq_labels}

\begin{figure}[!htb]
    \centering
    \includegraphics[width=\linewidth]{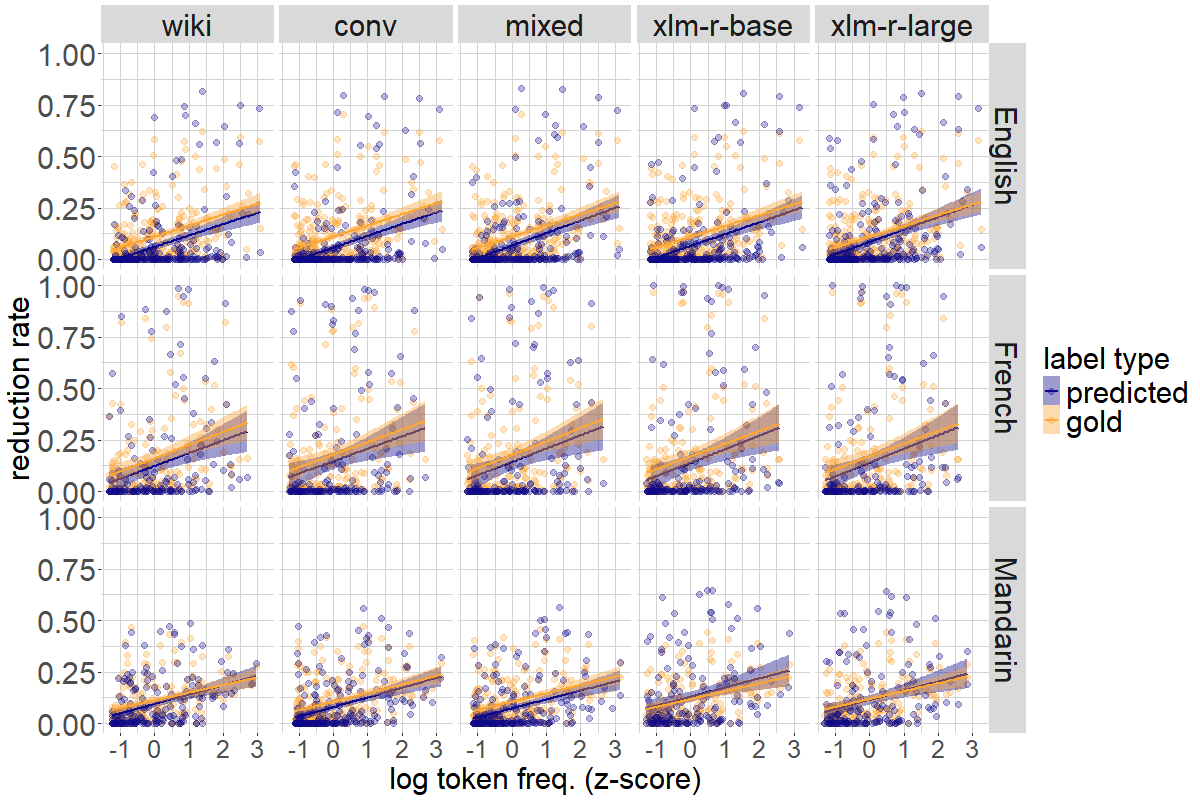}
    \caption{Predicted and gold reduction label rates as a function of log frequency (x axis, z-scored) and model type $\times$ language (panels)}
    \label{fig:freq_reduc}
\end{figure}

\begin{figure}[!htb]
    \centering
    \includegraphics[width=\linewidth]{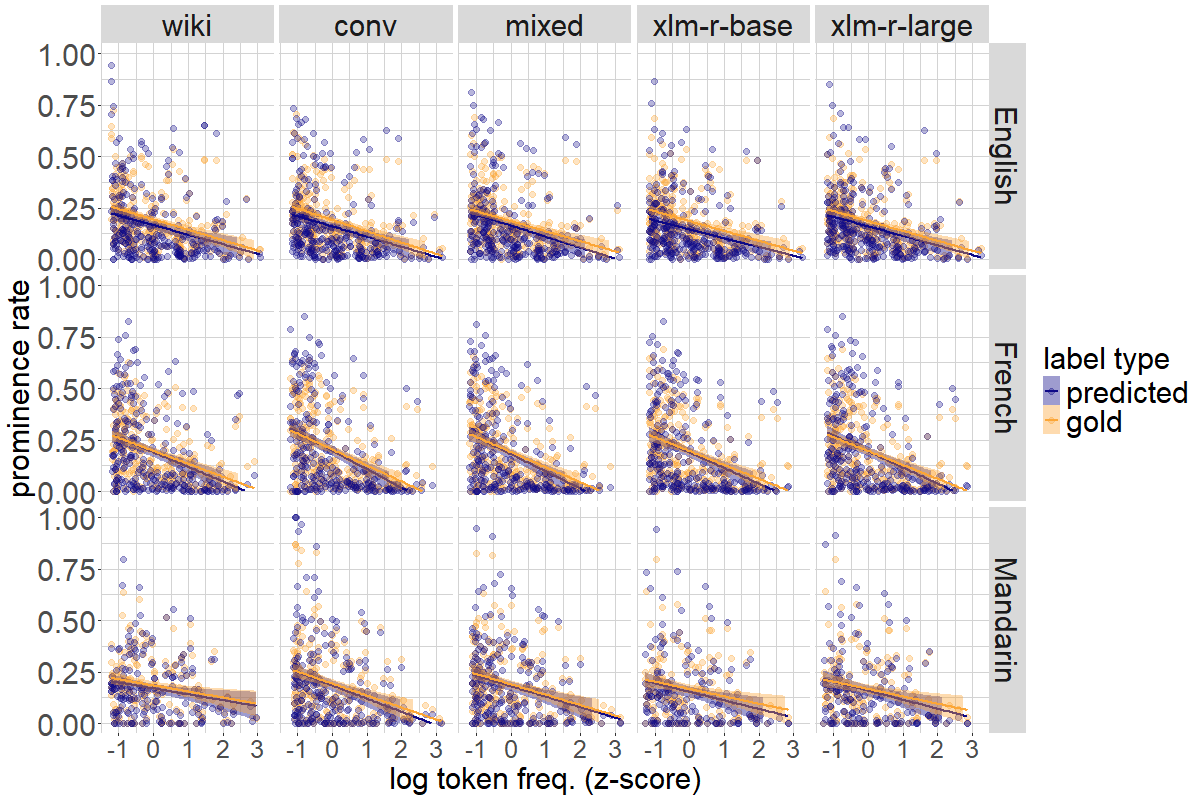}
    \caption{Predicted and gold prominence label rates as a function of log frequency (x axis, z-scored) and model type $\times$ language (panels)}
    \label{fig:freq_prom}
\end{figure}



\end{document}